\documentclass[11pt]{article}
\usepackage{eacl2017}
\usepackage{times}
\usepackage{url}
\usepackage{latexsym}

\usepackage{graphicx}
\usepackage[table]{xcolor}
\usepackage{array}
\usepackage{multirow}
\usepackage{tabularx}
\usepackage{multicol}
\usepackage{multirow}
\usepackage{diagbox}
\usepackage{amsmath}
\usepackage{amssymb}
\usepackage{booktabs}
\usepackage{float}
\restylefloat{table}
\usepackage[font=footnotesize,labelfont=bf]{caption}
\usepackage{enumitem}
\usepackage{makecell}

\newcolumntype{x}[1]{>{\centering\arraybackslash}p{#1}}

\usepackage[draft]{hyperref}
\hypersetup{
  hidelinks
}

\eaclfinalcopy

\title{Automatic Classification of Sentences in Paper Abstracts\\ with Neural Networks}
\title{Classification of Sentences in Paper Abstracts with Neural Networks}
\title{Sentence Classification in Medical Paper Abstracts with Neural Networks}
\title{Sentence Classification with Structured Prediction using Neural Networks}
\title{Neural Networks for Sequential Sentence Classification \\in Evidence-Based Medicine}
\title{Neural Networks for Joint Sentence Classification \\in Evidence-Based Medicine}
\title{Neural Networks for Joint Sentence Classification \\in Medical Paper Abstracts}

\author{Franck Dernoncourt\thanks{\hspace{3mm}These authors contributed equally to this work.}\\
	    MIT\\
	    {\tt francky@mit.edu}
	   \And
	 Ji Young Lee\footnotemark[1]\\
   	MIT\\
   {\tt jjylee@mit.edu}
   \And
	 Peter Szolovits \\
   	MIT\\
   {\tt psz@mit.edu}
   }
   
\date{}

\begin{document}
\maketitle
\begin{abstract}

Existing models based on artificial neural networks (ANNs) for sentence classification often do not incorporate the context in which sentences appear, and classify sentences individually. 
However, traditional sentence classification approaches have been shown to greatly benefit from jointly classifying subsequent sentences, such as with conditional random fields. In this work, we present an ANN architecture that combines the effectiveness of typical ANN models to classify sentences in isolation, with the strength of structured prediction. Our model achieves state-of-the-art results on two different datasets for sequential sentence classification in medical abstracts.

\end{abstract}

\section{Introduction}
Over 50 million scholarly articles have been published~\cite{jinha2010article}, and the number of articles published every year keeps increasing~\cite{druss2005growth,larsen2010rate}. Approximately half of them are biomedical papers.
While this repository of human knowledge abounds with useful information that may unlock new, promising research directions or provide conclusive evidence about phenomena, it has become increasingly difficult to take advantage of all available information due to its sheer amount.
Therefore, a technology that can assist a user to quickly locate the information of interest is highly desired, as it may reduce the time required to locate relevant information.

When researchers search for previous literature, for example, they often skim through abstracts in order to quickly check whether the papers match their criteria of interest. This process is easier when abstracts are \emph{structured}, i.e., the text in an abstract is divided into semantic headings such as objective, method, result, and conclusion. However, a significant portion of published paper abstracts is \emph{unstructured}, which makes it more difficult to quickly access the information of interest. Therefore, classifying each sentence of an abstract to an appropriate heading can significantly reduce time to locate the desired information.

We call this the \emph{sequential sentence classification task},
in order to distinguish it from general text classification or sentence classification that does not have any context. Besides aiding humans, this task may also be useful for automatic text summarization, information extraction, and information retrieval.

In this paper, we present a system based on ANNs for the sequential sentence classification task. 
Our model makes use of both token and character embeddings for classifying sentences, and has a sequence optimization layer that is learned jointly with other components of the model. We evaluate our model on the NICTA-PIBOSO dataset as well as a new dataset we compiled based on the PubMed database.

\section{Related Work}

Existing systems for sequential sentence classification are mostly based on naive Bayes (NB)~\cite{ruch2007using,huang2013pico}, support vector machines (SVMs)~\cite{mcknight2003categorization,yamamoto2005sentence,hirohata2008identifying}, Hidden Markov models (HMMs)~\cite{lin2006generative}, and conditional random fields (CRFs)~\cite{kim2011automatic,hassanzadeh2014identifying,hirohata2008identifying}. They often require numerous hand-engineered features based on lexical (bag-of-words, n-grams, dictionaries, cue words), semantic (synonyms, hyponyms), structural (part-of-speech tags, headings), and sequential (sentence position, surrounding features) information.

On the other hand, recent approaches to natural language processing (NLP) based on artificial neural networks (ANNs) do not require manual features, as they are trained to automatically learn features based on word as well as character embeddings. Moreover, ANN-based models have achieved state-of-the-art results on various NLP tasks. 
For short-text classification, many ANN models use word embeddings~\cite{socher2013recursive,kim2014convolutional,kalchbrenner2014convolutional}, and most recent works are based on character embeddings~\cite{zhang2015character,conneau2016very,xiao2016efficient}. Dos Santos and Gatti~\shortcite{dos2014deep} use both word and character embeddings.

However, most existing works using ANNs for short-text classification do not use any context. This is in contrast with \emph{sequential} sentence classification, where each sentence in a text is classified taking into account its context. The context utilized for the classification could be the surrounding sentences or possibly the whole text. 
One exception is a recent work on dialog act classification~\cite{lee2016sequential}, where each utterance in a dialog is classified into its dialog act, but only the preceding utterances were used, as the system was designed with real-time applications in mind.

\section{Model}

In the following, we denote scalars in italic lowercase (e.g., $k$, $b_f$), vectors in bold lowercase (e.g., $\mathbf{s},\, \mathbf{x}_i$), and matrices in italic uppercase (e.g., $W_f$) symbols.
We use the colon notations $x_{i:j}$ and $ \mathbf{v}_{i:j}$ to denote the sequences of scalars $(x_i, x_{i+1}, \dotsc, x_j)$ and vectors $(\mathbf{v}_i, \mathbf{v}_{i+1}, \dotsc, \mathbf{v}_j)$, respectively.

\subsection{ANN model}
\begin{figure}[!ht]
  \centering
  \includegraphics[width=0.45\textwidth]{{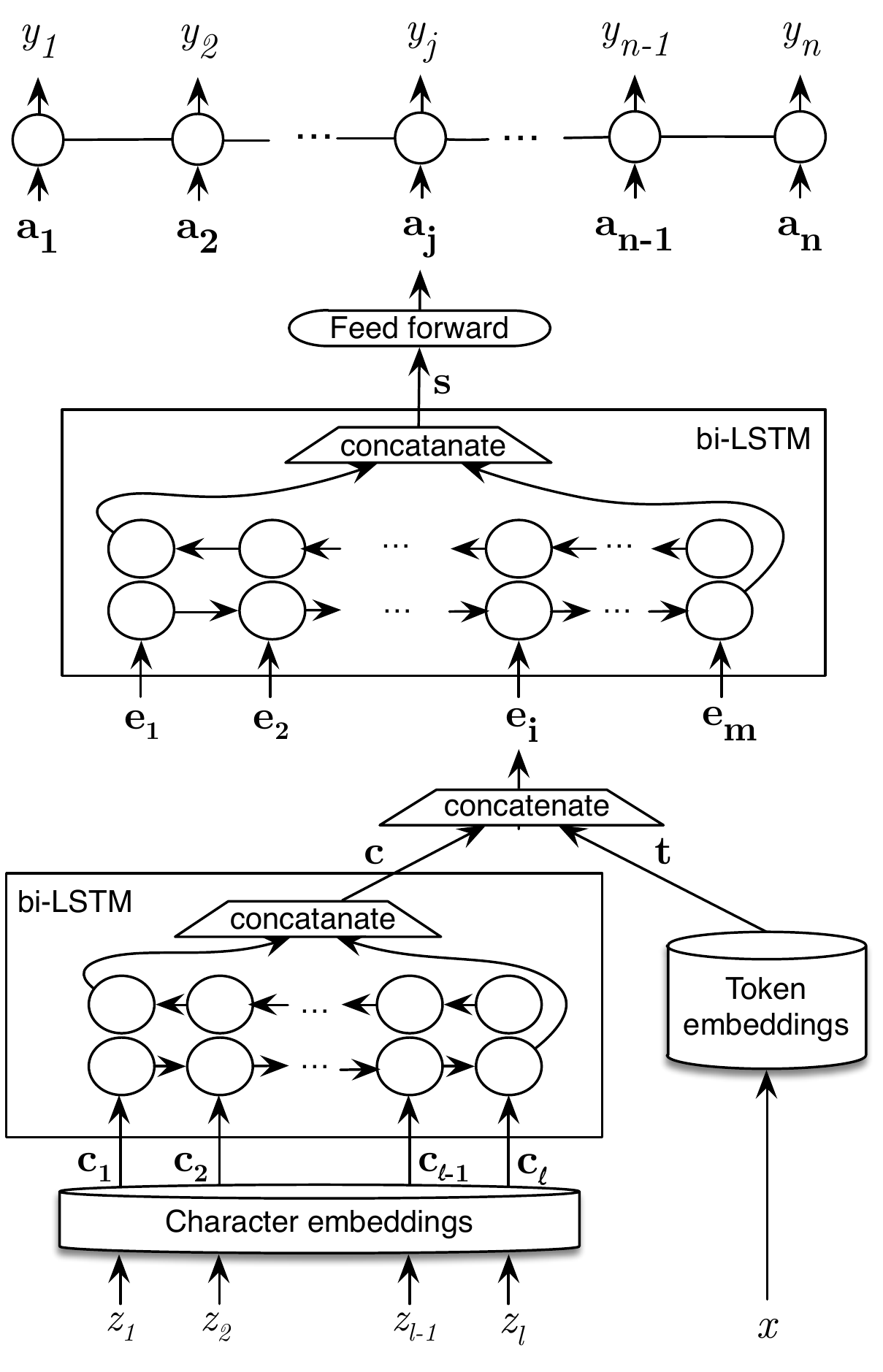}}
  \vspace{-0.4cm}
  \caption{ANN model for sequential sentence classification. $x$: token, $\mathbf{t}$: token embeddings (300), $z_i$: $i^{\text{th}}$ character of $x$, $\mathbf{c}_i$: character embeddings (25), $\mathbf{c}$: character-based token embeddings (50), $\mathbf{e}_i$: hybrid token embeddings (350), $\mathbf{s}$: sentence vector (200), $\mathbf{a}_j$: sentence label vector
  (number of classes), $y_j$: sentence label. The numbers in parenthesis indicate the dimensions of the vectors. Token embeddings are initialized with GloVe~\protect\cite{pennington2014glove} embeddings pretrained on Wikipedia and Gigaword 5~\protect\cite{parker2011english}. }
  \label{fig:model}
\end{figure}

Our ANN model (Figure~\ref{fig:model}) consists of three components: 
a hybrid token embedding layer, a sentence label prediction layer, and a label sequence optimization layer.

\subsubsection{Hybrid token embedding layer} \label{sec:embedding}

The hybrid token embedding layer takes a token as an input and outputs its vector representation utilizing both the token embeddings and as well as the character embeddings.

Token
embeddings are a direct mapping $\mathcal{V}_T(\cdot)$ from token to vector, which can be pre-trained on large unlabeled datasets using programs such as word2vec~\cite{mikolov2013distributed,mikolov2013efficient,mikolov2013linguistic} or GloVe~\cite{pennington2014glove}. 
Character embeddings are also defined in an analogous manner, as a direct mapping $\mathcal{V}_C(\cdot)$ from character to vector.

Let $z_{1:\ell}$
be the sequence of characters that comprise a token $x.$
Each character $z_i$ is first mapped to its embedding $\mathbf{c}_i = \mathcal{V}_C(z_i),$ and the resulting sequence
$\mathbf{c}_{1:\ell}$ is input to a bidirectional LSTM, which outputs the character-based token embedding $\mathbf{c}.$

The output $\mathbf{e}$ of the hybrid token embedding layer for the token $x$ is the concatenation of the character-based token embedding $\mathbf{c}$ and the token embedding $\mathbf{t} = \mathcal{V}_T(x).$

\subsubsection{Sentence label prediction layer} \label{sec:prediction}

Let $x_{1:m}$ be the sequence of tokens in a given sentence, and $\mathbf{e}_{1:m}$ be the corresponding embedding output from the hybrid token embedding layer. The sentence label prediction layer takes as input the sequence of vectors $\mathbf{e}_{1:m}$, and outputs $\mathbf{a}$, where the $k^{th}$ element of $\mathbf{a},$ denoted $\mathbf{a}[k],$ reflects the probability that the given sentence has  label $k$. 

To achieve this, the sequence $\mathbf{e}_{1:m}$ is first input to a bidirectional LSTM, which outputs the vector representation $\mathbf{s}$ of the given sentence. The vector $\mathbf{s}$ is subsequently input to a feedforward neural network with one hidden layer,
which outputs the corresponding probability vector $\mathbf{a}$.

\subsubsection{Label sequence optimization layer} \label{sec:sequence-optimization}

The label sequence optimization layer takes the sequence of probability vectors $\mathbf{a}_{1:n}$ from the label prediction layer as input, and outputs a sequence of labels $y_{1:n}$, where $y_{i}$ is the label assigned to the token $x_{i}$.

In order to model dependencies between subsequent labels, we incorporate a matrix $T$ that contains the transition probabilities between two subsequent labels; we define $T[i,j]$ as the probability that a token with label $i$ is followed by a token with the label $j$. The score of a label sequence $y_{1:n}$ is defined as the sum of the probabilities of individual labels and the transition probabilities:
$$ s(y_{1:n}) = { \sum_{i=1}^{n} \mathbf{a}_{i}[y_{i}]+  \sum_{i=2}^{n} T [y_{i-1},y_{i}} ]. $$
These scores can be turned into probabilities of the label sequences by taking a softmax function over all possible label sequences. 
 During the training phase, the objective is to maximize the log probability of the gold label sequence. In the testing phase, given an input sequence of tokens, the corresponding sequence of predicted labels is chosen as the one that maximizes the score. 

\vspace{0mm}
\section{Experiments}
\subsection{Datasets}

We evaluate our model on the sentence classification task using the following two medical abstract datasets, where each sentence of the abstract is annotated with one label. Table~\ref{tab:datasets} presents statistics on each dataset.

\paragraph{NICTA-PIBOSO}This dataset was introduced in~\cite{kim2011automatic} and was the basis of the ALTA 2012 Shared Task~\cite{amini2012overview}.
\paragraph{PubMed 20k RCT}We assembled this corpus consisting of randomized controlled trials (RCTs) from the PubMed database of biomedical literature, which provides a standard set of 5 sentence labels: objectives, background, methods, results and conclusions.

\begin{table} [H]
\footnotesize
\centering
\setlength\tabcolsep{4.0pt}
\setlength{\extrarowheight}{3pt}
\setlength{\arraycolsep}{5pt}
\begin{tabular}{|l|c|c|c|c|c|}
\hline
\textbf{Dataset} & \textbf{$|C|$} 	& \textbf{$|V|$} 	& Train & Validation & Test \\
\hline
\text{PubMed}	& 5	&	68k	& 15k (195k)	& 2.5k (33k) 	& 2.5k	(33k)	\\
\text{NICTA}		& 6		&	17k &	722 (8k)		& 77 (0.9k) 	& 200 (2k)	 \\
\hline
\end{tabular}
\caption{Dataset overview.   
$|C|$ denotes the number of classes, $|V|$ the vocabulary size. For the train, validation and test sets, we indicate the number of number of abstracts followed by the number of sentences in parentheses.
} \label{tab:datasets}
\end{table}

\vspace{5mm}
\subsection{Training}

The model is trained using stochastic gradient descent, updating all parameters, i.e., token embeddings, character embeddings, parameters of bidirectional LSTMs, and transition probabilities, at each gradient step. For regularization, dropout with a rate of $0.5$ is applied to the character-enhanced token embeddings and before the label prediction layer. 

\vspace{5mm}

\section{Results and Discussion}

Table~\ref{tab:result-comparisons} compares our model against several baselines as well as the best performing model~\cite{vrl2012feature} in the ALTA 2012 Shared Task, in which 8~competing research teams participated to build the most accurate classifier for the NICTA-PIBOSO corpus.

The first baseline (LR) is a classifier based on logistic regression using n-gram features extracted from the current sentence: it does not use any information from the surrounding sentences. The second baseline (Forward ANN) uses the model presented in~\cite{lee2016sequential}: it computes sentence embeddings for each sentence, then classifies the current sentence given a few preceding sentence embeddings as well as the current sentence embedding. The third baseline (CRF) is a CRF that uses n-grams as features: each output variable of the CRF corresponds to a label for a sentence, and the sequence the CRF considers is the entire abstract.
The CRF baseline therefore uses both preceding and succeeding sentences when classifying the current sentence. Lastly, the model presented in~\cite{vrl2012feature} developed a new approach called feature stacking, which is a metalearner that combines multiple feature sets, and is the best performing system on NICTA-PIBOSO published in the literature. 

\begin{table} [t]
\footnotesize
\centering
\setlength{\extrarowheight}{3pt}
\setlength{\arraycolsep}{5pt}
\begin{tabular}{|l|c|c|c|c|c|c|}
\hline
\textbf{Model} & PubMed 20k	& NICTA \\
 \hline

LR			& 83.0		&	71.6	 \\ 
Forward ANN		& 86.1			&	75.1	 \\ 
CRF		& 89.3			&	81.2	 \\
Best published		& --			&	82.0	 \\
Our model		& \textbf{89.9}			&	\textbf{82.7}	 \\
\hline
\end{tabular}
\caption{F1-scores on the test set with several baselines, the best published method \protect\cite{vrl2012feature} from the literature, and our model.
Since PubMed 20k was introduced in this work, there is no previous best published method for this dataset.
The presented results for the ANN-based models are the F1-scores on the test set of the run with the highest F1-score on the validation set.\vspace{-0.1cm}
} \label{tab:result-comparisons}
\end{table}

The LR system performs honorably on PubMed 20k RCT (F1-score: 83.0), but quite poorly on NICTA-PIBOSO (F1-score: 71.6): this suggests that using the surrounding sentences may be more important in NICTA-PIBOSO than in PubMed 20k RCT.

The Forward ANN system performs better than the LR system, and worse than the CRF: this is unsurprising, as the Forward ANN system only uses the information from the preceding sentences but does not use any information from the succeeding sentences,  unlike the CRF.

Our model performs better than the CRF system and the~\cite{vrl2012feature} system. We hypothesize that the following four factors give an edge to our model.

\noindent\textbf{No human-engineered features}: Unlike most other systems, our model does not rely on any human-engineered features.\\
\noindent\textbf{No n-grams}: While other systems heavily rely on n-grams, our model maps each token to a token embedding, and feeds it as an input to an RNN.
This helps combat data scarcity: for example, ``chronic tendonitis'' and ``chronic tendinitis'' are two different bigrams, but their token embeddings should be very similar since they share the same meaning.\\ 
\noindent\textbf{Structured prediction}: The labels for all sentences in an abstract are predicted jointly, which improves the coherence between the predicted labels in a given abstract.\\
\noindent\textbf{Joint learning}: Our model learned the features and token embeddings jointly with the sequence optimization.

\begin{figure}[t]
  \centering
  \includegraphics[width=0.52\textwidth]{{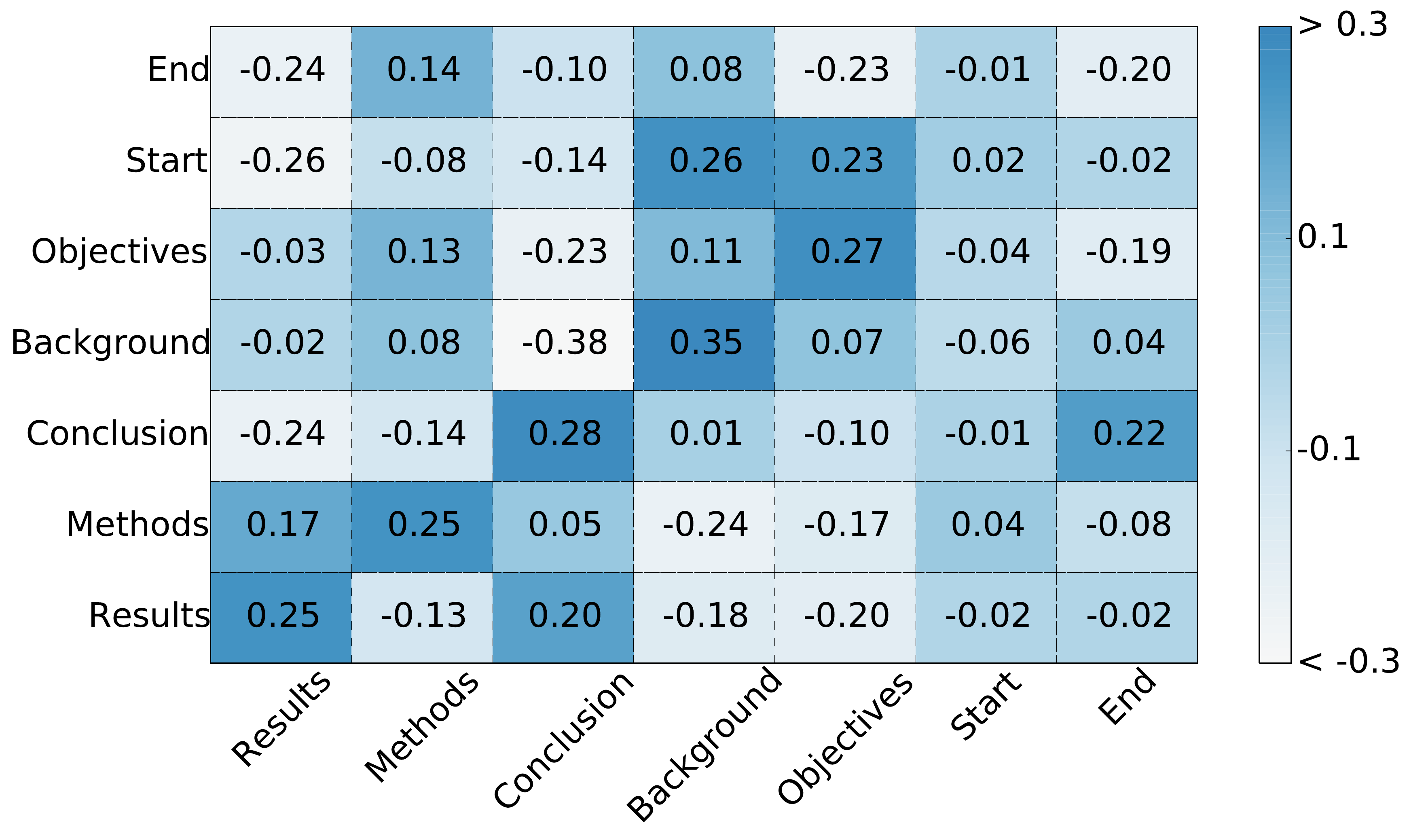}}
  \caption{Transition matrix learned on PubMed 20k. The rows represent the label  of the previous sentence, the columns represent the label of the current sentence.} 
  \label{tab:crf_transition_matrix}
\end{figure}

Figure~\ref{tab:crf_transition_matrix} presents an example of a transition matrix after the model has been trained on PubMed 20k RCT. We can see that it effectively reflects transitions between different labels. For example, it learned that the first sentence of an abstract is most likely to be either discussing objective ($0.23$) or background ($0.26$). By the same token, a sentence pertaining to the methods is typically followed by a sentence pertaining to the methods ($0.25$) or the results ($0.17$).

Table~\ref{tab:result-details} details the result of our model for each label in PubMed 20k RCT: the main difficulty the classifier has is distinguishing background sentences from objective sentences.

 \begin{table}[t]

\footnotesize
\centering
\setlength\tabcolsep{6.0pt}
\setlength{\extrarowheight}{3pt}
\setlength{\arraycolsep}{5pt}
\begin{tabular}{|c|cccc|}
\hline 
\multirow{2}{*}{Label} & \multicolumn{4}{c|}{PubMed 20k RCT} \tabularnewline
\cline{2-5} 
 & Precision & Recall & F1-score & Support \tabularnewline
\hline 
Background	& 71.8	& 88.2	& 79.1 	& 3621 \tabularnewline
Conclusion	& 93.5	& 92.9	& 93.2	& 4571 \tabularnewline
Methods		& 93.7	& 96.2	& 94.9	& 9897 \tabularnewline
Objectives	& 78.2	& 48.1	& 59.6	& 2333 \tabularnewline
Results	 	& 94.8 	& 93.1	& 93.9	& 9713 \tabularnewline
Total	 	& 90.0	& 89.8	& 89.9 	& 30135\tabularnewline
\hline 
\end{tabular}
\caption{Detailed results of our model on the PubMed 20k RCT dataset.
} 
\label{tab:result-details}
\end{table}

\vspace{0mm}

\vspace{-0mm}
\section{Conclusions}

In this article we have presented an ANN architecture to classify sentences that appear in sequence. 
We demonstrate that jointly predicting the classes of all sentences in a given text improves
the quality of the predictions and yields better performance than a CRF.
Our model achieves state-of-the-art results on two datasets for sentence classification in medical abstracts.

\bibliography{eacl2017}
\bibliographystyle{eacl2017}

\appendix

\end{document}